\documentclass[runningheads]{llncs}
\usepackage{graphicx}
\usepackage[misc]{ifsym}

\usepackage{stackengine}
\usepackage[ruled,lined,noend]{algorithm2e}          %
\usepackage{amsmath}
\usepackage{bm}
\usepackage{algcompatible}
\usepackage{dblfloatfix}        
\newcommand\xrowht[2][0]{\addstackgap[.5\dimexpr#2\relax]{\vphantom{#1}}}
\usepackage{bbm}
\usepackage{multirow}

\usepackage{array}
\usepackage{caption}
\usepackage{float}

\newcommand{\name}{AGRA }

\begin{document}
\title{Learning with Noisy Labels by Adaptive Gradient-Based Outlier Removal}
\titlerunning{Learning with Noisy Labels by Adaptive Gradient-Based Outlier Removal}
%
\author{Anastasiia Sedova{$^{1,2\star}$}(\Letter) \and 
Lena Zellinger{$^{1\star}$} \and 
Benjamin Roth$^{1,3}$}
\authorrunning{Sedova et al.}


\institute{$^1$Research Group Data Mining and Machine Learning, University of Vienna, Austria \\ 
$^2$UniVie Doctoral School Computer Science, University of Vienna, Austria \\ 
$^3$Faculty of Philological and Cultural Studies, University of Vienna, Austria \\
\email{\{anastasiia.sedova, lena.zellinger, benjamin.roth\}@univie.ac.at}}

\toctitle{Learning with Noisy Labels \\ by Adaptive Gradient-Based Outlier Removal}
\tocauthor{Anastasiia~Sedova, Lena~Zellinger, Benjamin~Roth}

\maketitle              
\def\thefootnote{$\star$}\footnotetext{Equal contribution}\def\thefootnote{\arabic{footnote}}
\begin{abstract}
An accurate and substantial dataset is essential for training a reliable and well-performing model.
However, even manually annotated datasets contain label errors, not to mention automatically labeled ones.
Previous methods for label denoising have primarily focused on detecting outliers and their permanent removal -- a process that is likely to over- or underfilter the dataset.
In this work, we propose AGRA: a new method for learning with noisy labels by using Adaptive GRAdient-based outlier removal\footnote{We share our code at: https://github.com/anasedova/AGRA}. 
Instead of cleaning the dataset \textit{prior} to model training, the dataset is dynamically adjusted \textit{during} the training process.
By comparing the aggregated gradient of a batch of samples and an individual example gradient, our method dynamically decides whether a corresponding example is helpful for the model at this point or is counter-productive and should be left out for the current update.
Extensive evaluation on several datasets demonstrates AGRA's effectiveness, while a comprehensive results analysis supports our initial hypothesis: permanent hard outlier removal is not always what model benefits the most from.

\end{abstract}
\section{Introduction}

The quality and effectiveness of a trained model heavily depend on the  quality and quantity of the training data. 
However, ensuring \textit{consistent} quality in automatic or human annotations can be challenging, especially when those annotations are produced under resource constraints or for large amounts of data. 
As a result, real-world datasets often contain annotation errors, or ``label noise", which can harm the model's overall quality.

Previous data-cleaning methods for noise reduction have attempted to improve the data quality by identifying and removing ``noisy", i.e., mislabeled samples before model training.
Some approaches detect noisy samples based on the disagreement between assigned and predicted labels in a cross-validation setting \cite{northcutt2021confident,crossweigh}, while others leverage knowledge transferred from a teacher model trained on clean data \cite{8237473}.
Such approaches typically rely on certain assumptions regarding the label noise, for instance, that the noise follows some particular distribution, is symmetric \cite{DBLP:conf/iccv/HuangQJZ19,DBLP:conf/icml/ArazoOAOM19}, or class-conditioned \cite{northcutt2021confident}. 
However, the \emph{true} data-generating process and noise level are usually unknown, and these methods easily over- or under-filter the data.

Another subtle problem arises from the \textit{static} nature of these methods, as they do not address the cases when problematic training samples for one model actually be beneficial for another.
Take the hypothetical -- wrongly labeled -- movie review:\\

\begin{center}
    \emph{``The movie was by no means great.''} -- \texttt{POSITIVE}
    \newline
\end{center}
Despite the incorrect label, a model that does not know anything about sentiment prediction still might learn the useful association between the word \emph{great} and the class \texttt{POSITIVE}. 
Therefore, this sample could be a valuable contribution to the training process.
On the other hand, the same sample might be confusing and deteriorating for the same model at a different training stage, when it already learned to distinguish subtle language phenomena like negation.

\begin{figure*}[t!]
\centering
\includegraphics[scale=0.3]{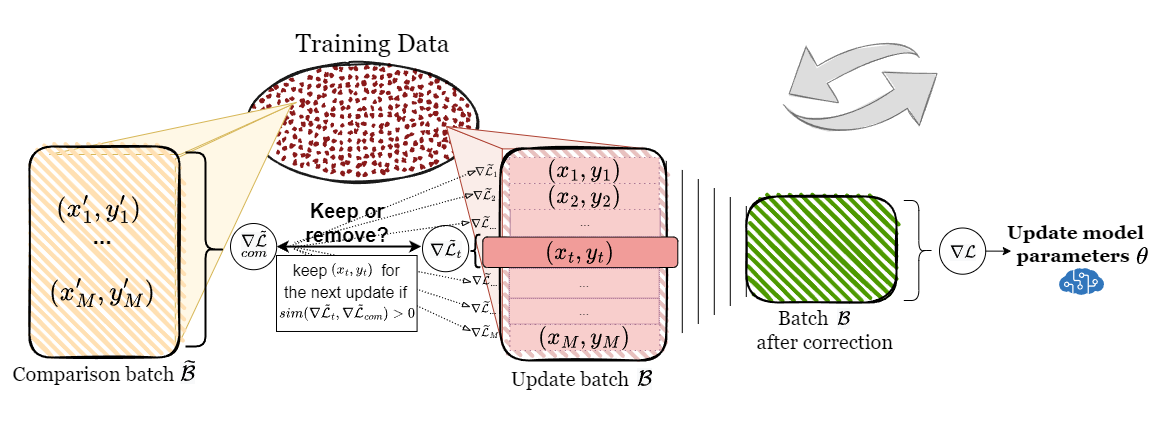}          
\caption{
\name method for learning with noisy data.
Each sample in the \textit{update batch} is decided to be either kept for further model training or removed depending on the similarity of its gradient to the aggregated gradient of the \textit{comparison batch} sampled from the same data.
}
\label{fig:agra}
\end{figure*}

In this paper, we reconsider the original motivation behind noise reduction:
instead of tracing the outliers, we focus on \textit{obtaining a model that remains unaffected by inconsistent and noisy samples.}
To achieve that, we suggest to dynamically adjust the training set during the training process instead of denoising it beforehand.
This idea is realized in our new method \textbf{\name} - a method for \textbf{A}daptive \textbf{GRA}dient-based outlier removal (see Figure~\ref{fig:agra}).
AGRA does not rely on some static, model-independent, implied properties but leverages gradients to measure the sample-specific impact on the current model.
During classifier training, \name decides for each sample whether it is \textit{useful} or not for a model at the current training stage by comparing its gradient with an accumulated gradient of another batch that is independently sampled from the same dataset.
Depending on the state of the classifier and the experimental setup, the sample is either used in the model update, excluded from the update, or assigned to an alternative label. 
Importantly, the effect of the sample may change in the next epoch, when the model state has changed.
Apart from that, we experimented with different loss functions and adapted an $F_1$-based loss function which optimizes the model directly towards the $F_1$ performance metric.
Extensive experiments demonstrate the effectiveness of our method and show that the \textit{correctness} of a training sample (as measured by manual annotation) is not the same as its \textit{usefulness} for the training process. 
\name reliably detects the latter in a trade-off with the former, which is crucial for the performance of the trained classifier.



Overall, our main contributions are the following:
\begin{itemize}
    \item We propose a new gradient-based method for adaptive outlier removal, AGRA, which dynamically identifies unusual and potentially harmful training samples during the learning process and corrects or removes them. 
    Since in our setting labeling errors are unknown at training time, \name uses the detrimental effect on the model w.r.t. to a comparison batch as a proxy to harmfulness.
    \item We analyze the effect of cross-entropy- and $F_1$-based loss functions for computing the compared gradients and show that utilizing the $F_1$ loss can improve performance on multiple datasets.
    \item We demonstrate the effectiveness of our method on several benchmark datasets where our method outperforms other denoising methods trained in an analogous evaluation setup.
\end{itemize}





\section{Related Work}

The high demand for large-scale labeled training data to train a stable classifier forces researchers and practitioners to look for more feasible solutions than relying on domain experts to annotate the data \cite{JMLR:v13:raykar12a,tratz-hovy-2010-taxonomy,ratner2020snorkel}.
The cost of such approaches is usually the annotation quality, and the resulting datasets often contain mislabeled samples.
Moreover, label noise can also be detected in expert annotations due to different factors in the data collection process \cite{russakovsky2015imagenet,frenay2013classification}. 
As a result, even widely-used datasets may contain incorrect annotations \cite{northcutt2021confident}, emphasizing the necessity for methods that enable the learning of reliable models despite the presence of label noise.


\paragraph{Learning with Noisy Labels.}
There are multiple general strategies for handling potential label noise. 
\textit{Data-cleaning} approaches separate the denoising process and the training of the final model: likely mislabeled samples are first identified and removed or corrected, and then the final model is trained on the cleaned dataset \cite{DBLP:conf/iccv/HuangQJZ19,DBLP:conf/icml/LiptonWS18}. 
The INCV algorithm \cite{chen2019understanding} iteratively estimates the joint distribution between the true labels and the noisy labels using out-of-sample model outputs obtained by cross-validation.
On the basis of the estimated joint distribution, the number of labeling errors is gauged and likely mislabeled samples are removed. 
Cleanlab \cite{northcutt2021confident} estimates the confident joint between true and noisy labels relying on the assumption of class-dependent noise.
Instead of defining a denoising system that would clean the data \textit{before} the training of the end model, AGRA joins the denoising and training into a single process where denoising happens \textit{during} the training of a single model. 
Moreover, \name does not make any assumption regarding the label noise distribution.
Other approaches, commonly referred to as \textit{model-centric},  focus on modifying the model architecture or the loss function to facilitate learning with noisy data. 
Wang et al. \cite{wang2019symmetric} add a noise-tolerant term to the cross-entropy loss, Ziyin et al. \cite{ziyin2020learning} propose a gambler's loss function, and
Sukhbaatar et al. \cite{sukhbaatar2014training} add an additional noise layer to convolutional neural networks. 
Other authors have explored more intricate training strategies for learning with noisy annotations:  
for instance,
Li, Socher, and Hoi \cite{li2020dividemix} leverage ensemble methods, and Li et al. \cite{li2019learning} exploit meta-learning techniques.
In contrast to these approaches, \name does not put any restrictions on the loss function and does not alter the model architecture.

\paragraph{Outlier Detection.}
Outlier detection is crucial in many real-world applications, such as fraud detection and health diagnosis \cite{wang2019progress}.
There are several general approaches for identifying outliers \cite{wang2019progress}: \emph{distance-based} methods consider a sample an outlier if it is far away from its nearest neighbors \cite{knox1998algorithms,ghoting2008fast}, \emph{density-based} approaches declare samples in low-density regions as outliers \cite{breunig2000lof,bai2016efficient}, \emph{clustering-based} strategies identify samples that are not associated with a large cluster \cite{elahi2008efficient,al2009effective}. 
AGRA defines outliers in terms of their utility at the current training step and aims at removing the ones that harm the current model.

\paragraph{Weak Supervision.}
To reduce the need for manual annotations, datasets can be labeled by automated processes, commonly referred to as \textit{weak supervision} \cite{DBLP:journals/corr/abs-2107-00934,DBLP:journals/corr/abs-2010-04582,DBLP:journals/corr/abs-2006-11747}.
In the weakly supervised setting, expert knowledge and intuition are formalized into a set of rules, or labeling functions \cite{ratner2020snorkel}, which annotate the training samples with weak, potentially noisy labels.
Various approaches to denoise the weakly supervised data include leveraging labeling functions aggregation techniques \cite{ratner2020snorkel,metal}, learning via user feedback and manual correction \cite{hedderich}, separately modeling labeling-function-specific and task-specific information in the latent space \cite{stephan-etal-2022-sepll}, or utilizing a small set of manually annotated data in addition to the weakly supervised samples \cite{DBLP:journals/corr/abs-2104-05514}. 
In contrast to these methods, \name is not restricted to the weakly supervised setting (although it can be applied for it, even if the labeling functions are not accessible);
instead, it is applicable to any dataset that contains noisy labels, regardless of the labeling process used.

\paragraph{Gradient-based Approaches.}
\name is based on gradient comparisons, which were studied before in different contexts \cite{zhao2020dataset,shi2021gradient,jason}.
For instance, Zhao et al. \cite{zhao2020dataset} explored gradient matching for generating artificial data points that represent a condensed version of the original dataset.
Unlike their approach, \name does not create any new data instances but adjusts how the already provided ones are used during training.
Shi et al. \cite{shi2021gradient} leverage gradient matching for domain generalization. AGRA, on the other hand, tackles a different problem and does not explicitly assume distribution shifts in the data.

\section{AGRA: Adaptive Gradient-Based Outlier Removal}
\label{sec:algorithm}

The main goal of \name is detecting the instances that would harm the model in the current training stage and filter them out or assign them to another class before the update.
In order to decide which samples are potentially harmful, the model gradients for each sample in the \textit{update batch} (i.e., the batch used during the training process for the model update) are compared one by one with an aggregated gradient from another batch sampled from the same data (\textit{comparison batch}).
Informally, such an aggregated \textit{comparison gradient} could be seen as an expected weight change on mostly clean data, assuming that the overall noise rate is not too high.
If the \textit{update gradient} of a sample from the update batch and the comparison gradient point in opposing directions, this could be an indication that the sample is harmful to the training process at this stage.
We refer to such samples as \emph{outliers} since they may have a negative impact on the current model update, even though they are not necessarily mislabeled. 
Each identified outlier is either removed from the update batch to prevent its influence on the weight update or reassigned to another class if doing so results in a higher, positive gradient similarity.



Unlike common denoising approaches that clean and fix the training dataset for the training process, \name does not make any decisions about removing or relabeling the samples \textit{before} training the model.
Instead, samples are relabeled or removed from the update batch \emph{on the fly}, based on the model's current state. 
Their participation in gradient update can therefore be reconsidered in later epochs.
If the model profits from an (even potentially mislabeled) sample during a particular training stage, this sample is kept.
However, the same sample may be removed during another stage where it would harm the training process.




\subsection{Notation}

We denote the training set by $\mathcal{X}=((x_1, y_1), ..., (x_T, y_T))$, where $y_t$ denotes a potentially noisy label  associated with the input $x_t$. Each $y_t$ corresponds to one out of $K$ classes $\{c_1, ..., c_K\}$. The task is to utilize $\mathcal{X}$ to learn a classifier $f\left(\cdot; \theta\right)$, parameterized by $\theta$, using an \emph{update loss function} $\mathcal{L}\left(x, y\right)$.
Additionally, we define a \emph{comparison loss function} $\widetilde\mathcal{L}\left(x, y\right)$ that is used for computing the compared gradients.
\name does not put any restrictions on the used loss functions; the update loss $\mathcal{L}\left(x, y\right)$ and the comparison loss $\widetilde\mathcal{L}\left(x, y\right)$ can differ.

\subsection{Algorithm Description}

\name consists of a single model training loop.
For each update batch $\mathcal{B}$,
another batch $\widetilde{\mathcal{B}}$ of the same size is independently sampled from the training dataset $\mathcal{X}$.
While $\mathcal{B}$ is leveraged to adjust the model weights during training, $\widetilde{\mathcal{B}}$ represents the comparison batch that is used to detect outliers. 

First, the batch-wise gradient on the comparison batch $\widetilde{\mathcal{B}}$ is computed with respect to the loss function $\widetilde{\mathcal{L}}$ and flattened into a vector, resulting in the comparison gradient $\nabla\widetilde\mathcal{L}_{com}$. Then, the gradient for each individual data point $(x_t, y_t) \in \mathcal{B}$ is calculated with respect to the loss $\widetilde{\mathcal{L}}$ and flattened, resulting in $\nabla\widetilde\mathcal{L}\left(x_t,y_t\right)$. Next, the pair-wise cosine similarity of each per-sample gradient with the comparison gradient is computed as given below\footnote{The subscript $x_t$ is omitted in the short-hand notation $sim_{y_t}$ for brevity.}.

\begin{equation}
\small
sim_{y_t}=sim\left(\nabla\widetilde\mathcal{L}{\left(x_t,y_t\right)}, \nabla\widetilde\mathcal{L}_{com}\right)=\frac{\nabla\widetilde\mathcal{L}\left(x_t,y_t\right) \cdot \nabla\widetilde\mathcal{L}_{com}}{||\nabla\widetilde\mathcal{L}\left(x_t,y_t\right)||_{2}\,||\nabla\widetilde\mathcal{L}_{com}||_{2}}               
\label{sim}
\end{equation}
In the following, $sim_{y_t}$ is referred to as the \emph{similarity score given label $y_t$}.

The next step can be realized in two different settings: 

\begin{enumerate}
     \item \textit{without an alternative label}: a data sample is removed from the update batch if its associated similarity score is non-positive and kept otherwise:
     \[
    \mathcal{B} \leftarrow 
    \begin{cases}
        \mathcal{B} \setminus \{\left(x_t, y_t\right)\},& \text{if } sim_{y_t} \leq 0\\
        \mathcal{B},              & \text{otherwise}
    \end{cases}
    \]
     
     
    \item \textit{with an alternative label}: in addition to the options of removing a training instance or retaining it with its original annotation, the instance can also be included in the update with the alternative label $y'$.
    If such an alternative label $y'$ is specified, the similarity $sim_{y'}=sim\left(\nabla\widetilde\mathcal{L}\left(x_t, y'\right), \nabla\widetilde\mathcal{L}_{com}\right)$ is additionally calculated using Eq. \ref{sim}. 

    Depending on the values of $sim_{y_t}$ and $sim_{y'}$, the sample is handled as follows: 
    \begin{itemize}{
    \item if the similarity score is non-positive given both $y_t$ and $y'$, the sample is removed from the batch,
    \item if the similarity score given label $y'$ is positive and higher than the similarity score given $y_t$, the original label $y_t$ is changed to $y'$,
    \item if the similarity score given label $y_t$ is positive and higher than or equal to the similarity score given $y'$, the original label $y_t$ is kept.
    \[
    \mathcal{B} \leftarrow 
    \begin{cases}
        \small
        \mathcal{B} \setminus \{\left(x_t, y_t\right)\}, &\text{if } sim_{y_t} \leq 0, sim_{y'} \leq 0 \\
        \mathcal{B} \setminus \{\left(x_t, y_t\right)\} \cup \{\left(x_t, y'\right)\},&\text{if } sim_{y'} > 0, sim_{y'} > sim_{y_t} \\
         \mathcal{B}, &\text{otherwise}
    \end{cases}
    \]
The decision regarding the choice of an alternative label and its sensibility depends on the characteristics and requirements of the specific dataset.
An intuitive approach is to use a negative class if it is present in the data (e.g., \textit{``no\_relation"} for relation extraction, 
or \textit{``non\_spam"} for spam detection).}
\end{itemize}
\end{enumerate}

\begin{algorithm}[t]
\textbf{Input:} training set $\mathcal{X}$, initial model $f\left(\cdot; \theta\right)$, number of epochs $E$, batch size $M$, (optionally: alternative label $y'$) \\
\textbf{Output}: trained model $f\left(\cdot; \theta^*\right)$\\
\For{epoch = 1,..., $E$}{
\For{batch $\mathcal{B}$}{
Sample a comparison batch $\widetilde{\mathcal{B}}$, $\widetilde{\mathcal{B}} \subset \mathcal{X}$, $|\widetilde{\mathcal{B}}| = M$\\
Compute $\nabla\widetilde\mathcal{L}_{com}$ on $\widetilde{\mathcal{B}}$\\
  \For{$\left(x_t, y_t \right) \in \mathcal{B}$}{
    Compute $\nabla\widetilde\mathcal{L}\left(x_t,y_t\right)$\\
    $sim_{y_t} = sim\left(\nabla\widetilde\mathcal{L}\left(x_t,y_t\right), \nabla\widetilde\mathcal{L}_{com}\right)$ (Eq. \ref{sim})  \\
  \eIf{an alternative label $y'$ is specified}{
    Compute $\nabla\widetilde\mathcal{L}(x_t, y')$ \\
    $sim_{y'} = sim\left(\nabla\widetilde\mathcal{L}(x_t, y'), \nabla\widetilde\mathcal{L}_{com}\right)$ (Eq. \ref{sim})  \\
    \If{$sim_{y_t} \leq 0$ and $sim_{y'} \leq 0$}{
    $\mathcal{B} \leftarrow \mathcal{B} \setminus \{\left(x_t, y_t\right)\}$
    }
    \If{$sim_{y'} > 0$ and $sim_{y'} > sim_{y_t}$}{
    $\mathcal{B} \leftarrow \mathcal{B} \setminus \{(x_t, y_t)\} \cup \left\{(x_t, y'\right)\}$
    }
    }
    {\If {$sim_{y_t} \leq 0$}{
    $\mathcal{B} \leftarrow \mathcal{B} \setminus \{\left(x_t, y_t\right)\}$
    }
    }
  }
  $\theta \leftarrow Optim\left(\theta, \mathcal{B}, \mathcal{L}\right)$
 }
 }
\caption{AGRA Algorithm for Single-Label Datasets}
\label{alg:alg_single}
\end{algorithm}

After each sample in $\mathcal{B}$ is considered for removal or correction, the model parameters are updated with respect to $\mathcal{L}$ and $\mathcal{B}$ before the processing of the next batch starts.
The method is summarized in Algorithm \ref{alg:alg_single}, and the graphical explanation is provided in Figure \ref{fig:agra}.

\subsection{Comparison Batch Sampling}
\label{sec:sampling}
Since the comparison gradient is an essential component of AGRA's outlier detection, it should be sampled in a way that does not disadvantage instances of any class.
For datasets with a fairly even class distribution, randomly selecting samples from the training data might be sufficient to get a well-balanced comparison batch.
However, when dealing with imbalanced datasets, this approach may result in an underrepresentation of rare classes in the comparison batch.
Consequently, the gradients of samples belonging to rare classes may not match the aggregated gradient computed almost exclusively on instances assigned to more common classes.

For such cases, \name provides \textit{class-weighted sampling} in order to ensure a large enough fraction of minority class instances in the comparison batch.
The weight for class $c_k$ is computed as the inverted number of occurrences of class $c_k$ in training set $\mathcal{X}$\footnote{In our implementation, these weights are passed to \texttt{WeightedRandomSampler} \cite{sampler} provided by the \texttt{torch} library. Note that \texttt{WeightedRandomSampler} does not assume that the weights sum up to 1.}: $$\frac{1}{{\sum\limits_{t=1}^{|\mathcal{X}|}\mathbbm{1}\left(y_{t}=c_k\right)}}$$
As a result, samples of common classes are generally less likely to be included in the comparison batch than those of rare classes, turning the resulting comparison batch into a well-formed representative of all classes.

\subsection{Selection of Comparison Loss Functions}
\label{sec:f1_loss}
\name does not imply any restrictions on the choice of the comparison loss function. 
For example, it can be combined with a standard \textbf{cross-entropy (CE)} loss function, which is suitable for both binary and multi-class classification problems, or \textbf{binary cross-entropy (BCE)}, which is commonly used in the multi-label setting. 
However, despite its effectiveness in many scenarios, the cross-entropy loss has been shown to exhibit overfitting on easy and under-learning on hard classes when confronted with noisy labels \cite{wang2019symmetric}.
Overall, cross-entropy losses can hardly be claimed robust to noise, making learning with noisy data even more challenging.

Aiming at reducing this effect, we adapted an \textbf{\boldsymbol{$F_{1}$} loss function} which directly represents the performance metric and aims to maximize the $F_{1}$ score.           
The $F_{1}$ loss function is similar to the standard $F_{1}$ score with one major difference: the predicted labels used for the calculation of true positives, false positives, and false negatives are replaced by the model outputs transformed into predicted probabilities by a suitable activation function.
This modification enables the $F_{1}$ score to become differentiable, making it compatible with gradient-based learning methods.
In contrast to previous research on leveraging the $F_{1}$ score as a loss function \cite{benedict2021sigmoidf1}, we investigate $F_1$ loss variants outside of the multi-label setting and gauge its efficacy in the presence of label noise.


For the multi-class single-label case, the $F_{1}$ loss is based on macro-$F_1$ metric:

$$\mathcal{L}_{F_{1_M}}\left(\mathcal{B}\right) = 1-\frac{1}{K}\sum_{k=1}^{K}\frac{2\widehat{tp}_k}{2\widehat{tp}_k + \widehat{fp}_k + \widehat{fn}_k + \epsilon}$$ 

where
\begin{align*}
\small
\widehat{tp}_{k} &= \sum_{t=1}^{M} \hat{y}_{t,k} \times \mathbbm{1}\left(y_{t}=c_k\right),\\
\widehat{fp}_{k} &= \sum_{t=1}^{M} \hat{y}_{t,k} \times \left(1-\mathbbm{1}\left(y_{t}=c_k\right)\right),\\
\widehat{fn}_{k} &= \sum_{t=1}^{M} \left(1-\hat{y}_{t,k}\right) \times \mathbbm{1}\left(y_{t}=c_k\right)\\
\end{align*}
and $\hat{y}_{t,k}$ denotes the predicted probability of class $k$ for sample $t$ after application of the softmax, $\times$ represents the element-wise product, and $\epsilon=1e-05$ in our experiments.
The $F_{1}$ loss for the multi-class multi-label setting is also based on the macro-$F_{1}$ score, while for the binary single-label setting, it is based on the $F_{1}$ score of the positive class.
The exact formulas for these variants are provided in Appendix B. 
Our experiments demonstrate that the $F_{1}$ loss function is beneficial as a comparison loss for some datasets compared to the classic cross-entropy loss.
However, we emphasize that its use is not mandatory for our algorithm: \name is compatible with any loss function. 

\section{Experiments}

In this section, we demonstrate the performance of our algorithm on several noisy datasets, compare it with various baselines, and analyze the obtained results. 

\subsection{Datasets}
\label{sec:datasets}


We evaluate our method \name on seven different datasets.
First, we choose three weakly supervised datasets from the Wrench \cite{wrench} benchmark:
(1) \textbf{YouTube} \cite{7424299} and  
(2) \textbf{SMS} \cite{spam_data,DBLP:conf/iclr/AwasthiGGS20} are spam detection datasets, and
(3) \textbf{TREC} \cite{li-roth-2002-learning,DBLP:conf/iclr/AwasthiGGS20} is a dataset for question classification.
The labeling functions used to obtain noisy annotations based on keywords, regular expressions, and heuristics are provided in previous work \cite{wrench,DBLP:conf/iclr/AwasthiGGS20}.
Next, we evaluate \name on two weakly supervised topic classification datasets in African languages, namely
(4) \textbf{Yorùb\'a} and 
(5) \textbf{Hausa} \cite{afr}; the keyword-based labeling functions were provided by the datasets' authors.
In order to obtain noisy labels for the training instances of the above datasets, we apply the provided labeling functions and use simple majority voting with randomly broken ties.
Samples without any rule matches (which are 12\% in (1), 59\% in (2), 5\% in (3), and none in (4) and (5)) are assigned to a random class. 

Apart from NLP datasets, we also conduct experiments on two image datasets:
(6) \textbf{CIFAR-10} \cite{cifar}, for which the noisy labels were generated by randomly flipping the clean labels following Northcutt et al. \cite{northcutt2021confident} with 20\% noise and 0.6 sparsity,
and (7) \textbf{CheXpert}, a multi-label medical imaging dataset \cite{irvin2019chexpert}. 
Since CheXpert test set is not revealed in the interest of the CheXpert competition,
the original hand-labeled validation set was used as a test set as in previous works \cite{giacomello2021image}, while a part of the training set was kept for validation purposes.
We used the noisy training annotations provided by Irvin et al. \cite{irvin2019chexpert}, which were obtained by applying the CheXpert labeler to the radiology reports associated with the images\footnote{The reports are not publicly accessible; only the noisy labels are available for the training data. The gold labels are not provided.}.
Since CheXpert is a multi-label classification task, we adapt our algorithm to the multi-label setting by performing the gradient comparison with respect to each output node, allowing to ignore individual entries of the label vector. The exact algorithm can be found in Appendix B. 

The dataset statistics are collected in Table \ref{table:dataset_stats}.
Each dataset's noise amount was calculated by comparing the noisy training labels to the gold labels.
The gold training labels are not provided for CheXpert, so its noise rate value is missing in the table.
More details about dataset preprocessing, as well as the label distributions of the datasets, are provided in Appendix C.

\begin{table}[t!]
\centering
\begin{tabular}{lrrrrrr}
\hline
 Dataset &  \#Class &   \#Train & \#Dev &  \#Test & \%Noise \\
\hline
 YouTube &          2 &      1586 &    120 &     250 &     18.8\\
     SMS &          2 &      4571 &    500 &     500 &      31.9\\
    TREC &          6 &      4965 &    500 &     500 &      48.2\\
Yorùb\'a &          7 &      1340 &    189 &     379 &      42.3\\
   Hausa &          5 &      2045 &    290 &     582 &      50.6\\
CheXpert &         12 &    200599 &  22815 &     234 &      -\\
CIFAR-10 &         10 &  50000   & 5000  & 5000   &  20 \\
\hline
\end{tabular}
\caption{Datasets statistics. 
The percentage of noise is calculated by comparing the noisy labels to the gold-standard annotations.
}
\label{table:dataset_stats}
\end{table}

\subsection{Baselines}

We compare AGRA towards seven baselines.
For datasets that include gold training labels (i.e., all datasets in our experiments except CheXpert), we trained a (1) \textbf{Gold} model with ground-truth labels; it can serve as an upper-bound baseline.
(2) \textbf{No Denoising} baseline entails simple model training with the noisy labels, without any additional data improvement. 
(3) \textbf{DP} \cite{ratner2016data} stands for the Data Programming algorithm, which improves the imperfect annotations by learning the structure within the labels and rules in an unsupervised fashion by a generative model.
(4) \textbf{MeTaL} \cite{metal} combines signals from multiple weak rules and trains a hierarchical multi-task network.
(5) \textbf{FlyingSquid} \cite{pmlr-v119-fu20a} rectifies the noisy annotations using an Ising model by a triplet formulation. 
The experiments with the above baselines were realized using the Wrench framework \cite{wrench}.
In addition to the methods (3), (4), and (5) that are specifically designed for the weakly supervised setting, we also compare \name with two baselines
that have broader applicability for learning with noisy labels:
(6) \textbf{CORES$^2$} \cite{cheng2020learning}, which utilizes confidence regularization to sieve out samples with corrupted labels during training, and (7) \textbf{Cleanlab} \cite{northcutt2021confident}, which aims at detecting noisy annotations by estimating the joint distribution between noisy and true labels using the out-of-sample predicted probabilities. 

Since DP, MeTaL, and FlyingSquid require access to annotation rules and rule matches, they cannot be applied to non-weakly supervised datasets or other datasets for which this information is not available (such as CheXpert, for which the reports used for annotation are not publicly released).
In contrast, Cleanlab, CORES$^2$, and AGRA directly utilize noisy labels and do not require additional information regarding the annotations, making them more broadly applicable. 

\subsection{Experimental Setup}
\name was implemented based on Python using the PyTorch library.
We evaluate our method with a logistic regression classifier optimized with Adam\footnote{AGRA can also be used with any PyTorch-compatible deep model as our method has no model-related limitations.}.
For text-based datasets, we use a TF-IDF feature vectors to represent the data.
The CheXpert images were encoded with a fine-tuned EfficientNet-B0 \cite{tan2019efficientnet}, and the CIFAR-10 images were encoded with a fine-tuned ResNet-50 \cite{resnet} (following previous work \cite{northcutt2021confident}). 
More details on the data encoding and resulting feature vectors are provided in Appendix D.
In our experiments with TF-IDF representations, we found that the gradient entries corresponding to the biases of the model strongly influence the computed similarity scores despite being feature-independent. 
Hence, we exclude the elements corresponding to the biases when determining the gradient similarity for sparse features.
To make our experiments consistent, we apply the same strategy to CIFAR-10 and CheXpert.

\begin{table}[t!]
\centering
\begin{tabular}{lccccc|c|cc}
& \multicolumn{1}{p{1.2cm}}{\centering \textbf{YouTube} \\ \small{(Acc)}}
& \multicolumn{1}{p{1.2cm}}{\centering \textbf{SMS} \\ \small{(F1)}}
& \multicolumn{1}{p{1.2cm}}{\centering \textbf{TREC} \\ \small{(Acc)}}
& \multicolumn{1}{p{1.2cm}}{\centering \textbf{Yorùb\'a} \\ \small{(F1)} }
& \multicolumn{1}{p{1.2cm}|}{\centering \textbf{Hausa} \\ \small{(F1)}}
& {\centering \textbf{Avg.}}
& \multicolumn{1}{p{1.2cm}}{\centering \textbf{CIFAR} \\ \small{(Acc)}}
& \multicolumn{1}{p{1.2cm}}{\centering \textbf{CXT} \\ \small{(AUR)}}
\\
\hline
\hline 
Gold & 
$94.8$\scriptsize{$\pm 0.8$} & 
$95.4$\scriptsize{$\pm 1.0$} & 
$89.5$\scriptsize{$\pm 0.3$} & 
$57.3$\scriptsize{$\pm 0.4$} & 
$78.5$\scriptsize{$\pm 0.3$} & 
$83.1$ &
$83.6$\scriptsize{$\pm 0.0$} &
$-$\\
No Denoising & 
$87.4$\scriptsize{$\pm 2.7$} &
$71.7$\scriptsize{$\pm 1.4$} &
$58.7$\scriptsize{$\pm 0.5$} & 
$44.6$\scriptsize{$\pm 0.4$} & 
$39.7$\scriptsize{$\pm 0.8$} & 
$60.4$ &
$82.4$\scriptsize{$\pm 0.2$} &
$82.7$\scriptsize{$\pm 0.1$}
\\
\hline
\multicolumn{2}{p{2.8cm}}{\textit{Weak Supervision}} &&&&&&& \\
DP \small{\cite{ratner2016data}} & $90.8$\scriptsize{$\pm 1.0$} & 
$44.1$\scriptsize{$\pm 6.7$} & 
$54.3$\scriptsize{$\pm 0.5$} & 
$\textbf{47.8}$\scriptsize{$\pm 1.7$} & 
$40.9$\scriptsize{$\pm 0.6$} & 
$55.6$ & $-$ & $-$ \\
MeTaL \small{\cite{metal}} & 
$92.0$\scriptsize{$\pm 0.8$} & 
$18.3$\scriptsize{$\pm 7.8$} & 
$50.4$\scriptsize{$\pm 1.7$} & 
$38.9$\scriptsize{$\pm 3.1$} & 
$45.5$\scriptsize{$\pm 1.1$} & 
$49.0$ & $-$ & $-$ \\
FS \small{\cite{pmlr-v119-fu20a}} & 
$84.8$\scriptsize{$\pm 1.2$} & 
$16.3$\scriptsize{$\pm 6.0$} & 
$27.2$\scriptsize{$\pm 0.1$} & 
$31.9$\scriptsize{$\pm 0.7$} & 
$37.6$\scriptsize{$\pm 1.0$} & 
$39.6$ & $-$ & $-$ \\ 
\hline
\multicolumn{2}{p{2.8cm}}{\footnotesize{\textit{Noisy Learning}}} &&&&&&\\
CORES$^2$ \cite{cheng2020learning} & $88.8$\scriptsize{$\pm 3.6$} & $85.8$\scriptsize{$\pm 1.8$} & $61.8$\scriptsize{$\pm 0.5$} &
$43.0$\scriptsize{$\pm 0.7$} &
$\textbf{51.2}$\scriptsize{$\pm 0.5$} &
$66.1$&
$83.4$\scriptsize{$\pm 0.1$}&
$-$ \\
Cleanlab \small{\cite{northcutt2021confident}} &
$91.3$\scriptsize{$\pm 1.2$} &
$80.6$\scriptsize{$\pm 0.3$} &
$60.9$\scriptsize{$\pm 0.4$} &
$43.8$\scriptsize{$\pm 1.3$} &
$40.3$\scriptsize{$\pm 0.3$} &
$63.4$ &
$83.3$\scriptsize{$\pm 0.0$} &
$81.2$\scriptsize{$\pm 0.2$} \\
\hline
\textbf{\name} & 
$\textbf{93.9}$\scriptsize{$\pm 0.7$} & 
$\textbf{87.7}$\scriptsize{$\pm 1.2$} & 
$\textbf{63.6}$\scriptsize{$\pm 0.7$} & 
$46.9$\scriptsize{$\pm 1.5$} & 
$46.2$\scriptsize{$\pm 1.6$} &
$\textbf{67.7}$ &
$\textbf{83.6}$\scriptsize{$\pm 0.0$} &
$\textbf{83.9}$\scriptsize{$\pm 0.3$}
\end{tabular}
\caption{Experimental results on NLP and image datasets averaged across five runs and reported with standard deviation.
}
\label{tab:results}
\end{table}

For each dataset, we report the same evaluation metrics as in previous works: commonly used accuracy and $F_1$ scores and macro-AUROC (Area Under the Receiver Operating Characteristics) for CheXpert \cite{irvin2019chexpert}\footnote{The AUROC was computed on the nine classes which have more than one positive observation in the test set.}.
The hyper-parameters were selected with a grid search; more details and the selected parameter values are provided in Appendix E\footnote{The experiments are performed on a single Tesla V100 GPU on Nvidia DGX-1.
The parameter grid search takes between 24 and 96 hours; the model run with the selected parameters takes between 1 and 5 minutes depending on parameter values and dataset size.}. 
After training each model for 10 epochs (5 epochs for CheXpert), we reload the best model state based on validation performance and evaluate it on the test set.

\subsection{Results}

The results of the experiments across the seven datasets are summarized in Table~\ref{tab:results}.
\name is the best-performing method overall for three weakly-supervised NLP datasets, providing better results than the methods specifically designed for weakly supervised data. 
Among the text-based datasets, the average improvement achieved by \name over FlyingSquid, MeTaL, and DP is $28.1$ percentage points (pp), $18.7$pp, and $12.1$pp correspondingly. 
Compared to the baselines designed for weakly supervised data, Cleanlab and CORES$^2$ worked better on average, but \name also demonstrates an improvement over them (by $4.3$pp and $1.6$pp, respectively). 
Notably, \name improves the results of \textit{all} datasets over simple training without additional denoising ($7.3$pp improvement on average).
For image datasets, AGRA also demonstrates an improvement over Cleanlab and CORES$^2$ as well as the no denoising baseline; for Cifar dataset, our method demonstrates even the same result as the model trained with gold labels.
The other baselines are not applicable to these datasets\footnote{The weak supervision baselines cannot be run on CIFAR since it is a non-weakly supervised dataset; they also cannot be run on CheXpert as we do not have access to the labeling function matches. Furthermore, CORES$^2$ is not applicable for CheXpert as it does not support multi-label settings.}.

\begin{table}[t!]
\centering
\begin{tabular}{l|c|c|c|c}
&
\multicolumn{2}{c}{No Weighted Sampling}&
\multicolumn{2}{|c}{Weighted Sampling}
\\
\hline
\hline\xrowht[()]{5pt} &
\multicolumn{1}{p{1.5cm}|}{\centering CE/CE}&
\multicolumn{1}{p{1.5cm}|}{\centering CE/$F_{1}$}&
\multicolumn{1}{p{1.5cm}|}{\centering CE/CE}&
\multicolumn{1}{p{1.5cm}}{\centering CE/$F_{1}$}
\\
\hline
YouTube & 
$92.0 \pm 1.0$ & 
$\mathbf{93.9 \pm 0.7}$ & 
$91.9 \pm 0.5$ & 
$93.4 \pm 0.8$
\\ 
YouTube$^\dagger$ & 
$90.5 \pm 1.0$&
$-$&
$92.0 \pm 0.7$&
$-$
\\
SMS & 
$79.0 \pm 3.2$ & 
$61.1 \pm 5.2$ & 
$\mathbf{87.7 \pm 1.2}$ & 
$49.1 \pm 3.0$
\\
SMS$^\dagger$ & 
$71.1 \pm 3.1$ &
$-$ &
$86.3 \pm 1.2$ &
$-$
\\ 
TREC & 
$61.6 \pm 0.6$ & 
$62.1 \pm 0.4$ &
$62.8 \pm 1.1$ &
$\mathbf{63.6 \pm 0.7}$
\\
Yorùb\'a & 
$44.3 \pm 2.5$ & 
$44.2 \pm 1.4$ & 
$43.5 \pm 1.0$ & 
$\mathbf{46.9 \pm 1.5}$ 
\\
Hausa & 
$41.2 \pm 0.4$ & 
$40.9 \pm 0.6$ & 
$43.8 \pm 2.8$ & 
$\mathbf{46.2 \pm 1.6}$
\\
CheXpert & 
$ 82.6 \pm 0.6$ & 
$\mathbf{83.9 \pm 0.3}$ & 
$-$ & 
$-$
\\
CIFAR & 
$82.2 \pm 0.2$ &
$83.5 \pm 0.0$ &
$83.1 \pm 0.0$ &
$\mathbf{83.6 \pm 0.0}$
\
\end{tabular}
\caption{\name experimental test results with different settings: use of class-weighted sampling, [training loss]/[comparison loss]. 
The results marked with $\dagger$ are obtained by \name with an alternative label.
All results are averaged across 5 runs and reported with standard deviation.}
\label{tab:all_results}
\end{table} 

\subsection{Ablation Study}

Table \ref{tab:all_results} shows the \name performance across all comparison losses and comparison batch sampling strategies, the best of which was included in Table \ref{tab:results}. 
Overall, it outperforms the baselines on most of the datasets in the vast majority of settings. 
For the binary YouTube and SMS datasets, we also perform experiments with an alternative label (the negative \textit{``non\_spam"} class for both datasets). 
However, the models trained with the alternative label setting are not the best-performing AGRA configuration for either dataset (although they sometimes outperform the corresponding settings without the alternative label).
Utilizing an $F_1$-based comparison loss function instead of standard cross entropy proved beneficial on all datasets except SMS.

\begin{figure*}[h!]
\centering
\includegraphics[scale=0.255]{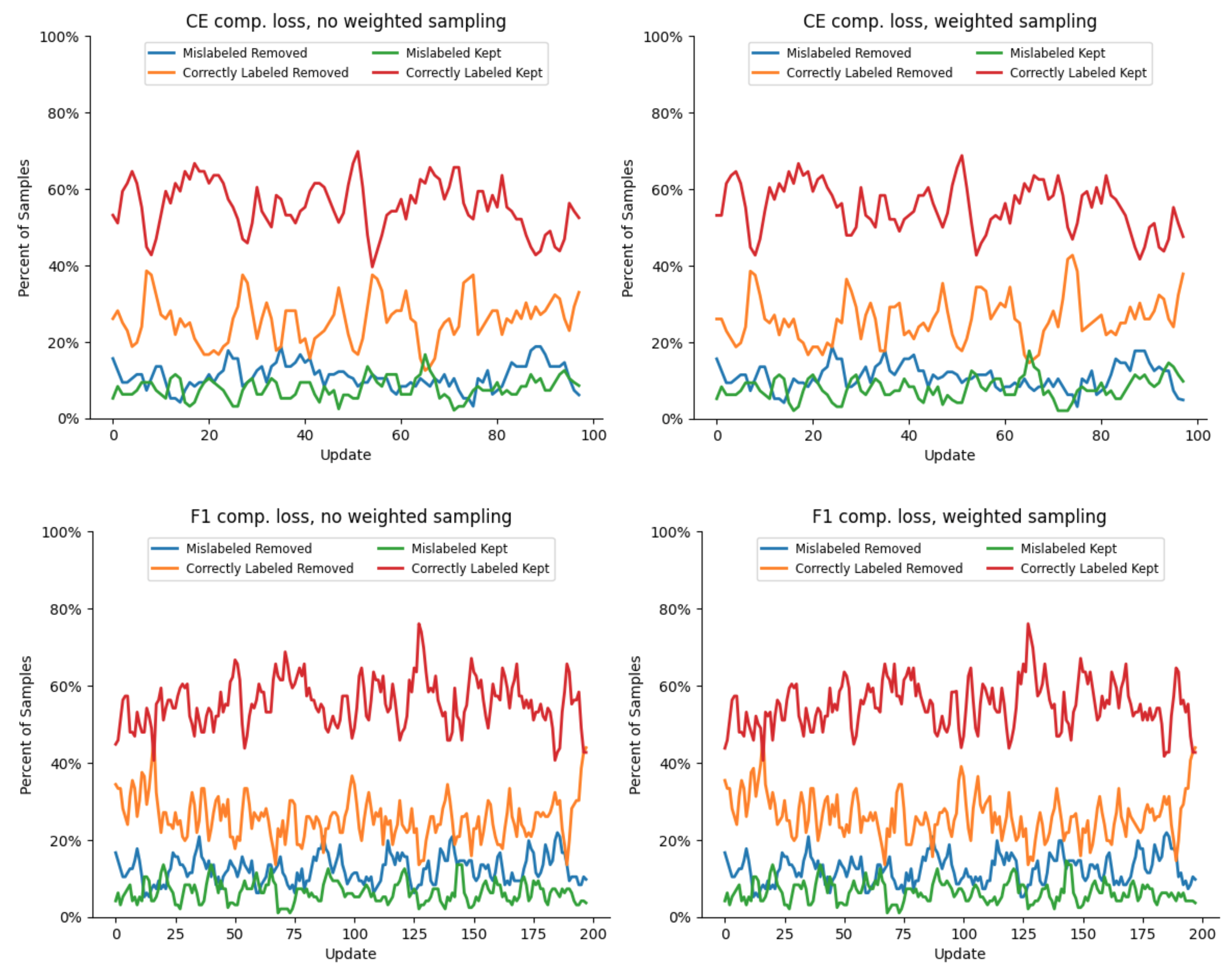}
\caption{Case study on the YouTube dataset. The plots represent the percentage of samples in each batch that were correctly kept, correctly removed, falsely kept and falsely removed during the training of the best-performing models for all combinations of comparison losses and sampling strategies.}
\label{fig:yoruba_gold}
\end{figure*}

As expected, weighted comparison batch sampling turns out to be especially helpful for imbalanced datasets such as Hausa (for which the most popular class is represented by $53.7 \%$ training samples, while the least frequent class only covers $7.9\%$) and TREC ($56.6\%$ and $1.0\%$ correspondingly; see detailed statistics in Appendix C).
On the other hand, the fairly balanced YouTube dataset performs marginally better without it.

\subsection{Case study}
\label{sec:case_study}

Finally, we provide a more fine-grained analysis of our \name method on the example of the YouTube dataset.
By comparing the noisy labels to the manual labels provided for this dataset, we calculate the fraction of samples in each batch that are (1) mislabeled and removed, (2) correctly labeled and removed, (3) mislabeled and kept, (4) correctly labeled and kept.
These statistics are reflected in Figure \ref{fig:yoruba_gold} for all supported combinations of comparison losses and comparison batch sampling strategies\footnote{The hyper-parameters chosen by the grid search were used for each depicted run. 
The exact values are provided in Appendix E.}.
A remarkable trend is that the correctness of removed samples appears to be not crucial for training a reliable model.
In fact, the amount of mislabeled samples kept and correctly labeled samples removed is high for many batches, yet all configurations outperform the baselines (excluding MeTaL which ties with the CE-based settings). 
This observation affirms our hypothesis that mislabeled samples can be beneficial at certain stages of the training process, and cleaning the dataset by filtering out all presumably mislabeled samples before training (as is done in common data-cleaning methods) might be a suboptimal approach.
Moreover, the number of ``falsely'' kept (\textit{mislabeled kept}) and removed (\textit{correctly labeled removed}) samples in some cases exceeds the amount of ``correctly'' kept (\textit{correctly labeled kept}) and removed (\textit{mislabeled removed}) ones.
This observation reinforces our point that the usefulness of a sample at the current training stage cannot be solely determined by whether it is mislabeled or not.
Weighted comparison batch sampling seems to only have a minor influence on the training process for YouTube. This observation can likely be explained by the already balanced noisy label distribution of the YouTube dataset.

\section{Conclusion}
In this work, we address the challenge of training a classifier using noisy labels.
Most importantly, we reconsider the goal of learning with noisy annotations and focus on training a stable and well-performing classifier rather than obtaining clean and error-free data. 
Instead of following the traditional approach of first denoising the data and then training a classifier on the cleaned data, we propose a novel integrated approach that dynamically adjusts the \emph{use} of the dataset during the learning process.
In our new algorithm AGRA, samples from which the model can benefit at the current training stage are retained for updating, while the ones that may hinder the learning process are disregarded or relabeled.
Our algorithm outperforms several recent baselines for training with noisy data.

\section{Ethical Statement}

Our method can improve the model predictions and produce more useful results, but we cannot promise they are perfect, especially for life-critical domains like healthcare. 
Data used for training can have biases that machine learning methods may pick up, and one needs to be careful when using such models in actual applications. 
We relied on datasets that were already published and did not hire anyone to annotate them for our work.

\section{Acknowledgement}
This research was funded by the WWTF through the project ”Knowledge-infused Deep Learning for Natural Language Processing” (WWTF Vienna Re- search Group VRG19-008).

\bibliographystyle{splncs04}
\bibliography{main}

\appendix

\section{$F_{1}$-Based Losses}

This section contains the formulas for the $F_1$-based loss functions used in the experiments.   
In the following, $\hat{y}_{t, k}$ denotes the model output for sample $t$ corresponding to the $k$-th class after application of an activation function.
In the binary single-label setting, where $y_{t} \in \{0, 1\}$, the positive class is assumed to be denoted by label 1. A small stabilizer $\epsilon$ was added to the denominator of the $F_1$ formula, inspired by past research on $F_1$-based loss functions\footnote{e.g., \url{https://towardsdatascience.com/the-unknown-benefits-of-using-a-soft-f1-loss-in-classification-systems-753902c0105d}, accessed: 2022-08-11}; $\epsilon=1e-05$ was used in the experiments.

\subsubsection{Binary-$F_1$ Loss for Single-Label Settings}
The $F_1$ loss geared towards binary classification tasks aims to maximize the $F_1$ score of the positive class. Given a batch $\mathcal{B}$, the loss can be computed by the following formula.
$$\mathcal{L}_{F_1}(\mathcal{B}) = 1-\frac{2\widehat{tp}}{2\widehat{tp} + \widehat{fp} + \widehat{fn} + \epsilon},$$ where

\begin{align*}
\widehat{tp} &= \sum_{t=1}^{M} \hat{y}_{t,1} \times y_{t},\\
\widehat{fp} &= \sum_{t=1}^{M} \hat{y}_{t,1} \times (1-y_{t}),\\
\widehat{fn} &= \sum_{t=1}^{M} (1-\hat{y}_{t,1}) \times y_{t},\\
\end{align*}
$y_t \in \{0,1\}$ and $\hat{y}_{t,1}$ denotes the predicted probability for the positive class for sample $t$ after application of the softmax.

\subsubsection{Macro-$F_1$ Loss for Multi-Label Settings} 
In the multi-label case, we use a macro-averaged $F_1$ loss, which averages the differentiable $F_1$ scores of all $K$ classes.

$$\mathcal{L}_{F_{1_M}}(\mathcal{B}) = 1-\frac{1}{K}\sum_{k=1}^{K}\frac{2\widehat{tp}_k}{2\widehat{tp}_k + \widehat{fp}_k + \widehat{fn}_k + \epsilon},$$ where

\begin{align*}
\widehat{tp}_{k} &= \sum_{t=1}^{M} \hat{y}_{t,k} \times y_{t,k},\\
\widehat{fp}_{k} &= \sum_{t=1}^{M} \hat{y}_{t,k} \times (1-y_{t,k}),\\
\widehat{fn}_{k} &= \sum_{t=1}^{M} (1-\hat{y}_{t,k}) \times y_{t,k},\\
\end{align*}
 $y_{t,k} \in \{0,1\}$ and $\hat{y}_{t,k}$ denotes the predicted probability of class $k$ for sample $t$ after application of the sigmoid function.




\section{AGRA Multi-Label Setting}
In the multi-label setting, the annotation for each data point is given by a binary vector of size $K$, where $K$ is the number of classes. The \name multi-label setting allows to ignore individual entries of this vector for the update.
For the following formulation of the multi-label variant of AGRA, the label-specific gradients are denoted by $\left(\nabla\widetilde\mathcal{L}\left(x_t,y_{t}\right)\right)_{k}$ and $\left(\nabla\widetilde\mathcal{L}_{com}\right)_{k}$, where $k \in \{1,...,K\}$, and the corresponding similarity score is given by $sim_{y_t}^k=sim\left(\left(\nabla\widetilde\mathcal{L}\left(x_t,y_t\right)\right)_{k}, \left(\nabla\widetilde\mathcal{L}_{com}\right)_{k}\right)$. 
If a particular label $y_{t,k}$ should not be considered for the update, due to $sim_{y_t}^k$ being non-positive, it is replaced with the label $i$. For the model update, we use a masked version of the BCE loss that ensures that the replaced labels do not influence the weight update. Namely,

$$\mathcal{\widetilde{L}}_{BCE}(\mathcal{B})=\frac{1}{K}\sum_{k=1}^{K}\frac{1}{\widetilde{M}_k}\sum_{t=1}^{M}-[\log(\hat{y}_{t,k})y_{t,k}+\log(1-\hat{y}_{t,k})(1-y_{t,k})]\mathbbm{1}(\widetilde{y}_{t,k}\neq i)$$

where $\widetilde{M}_k$ denotes the number of retained labels for class $k$ and $\widetilde{y}$ denotes the masked label vector.

Pseudocode for the \name multi-label setting can be found in Algorithm \ref{alg:alg_multi}.

\begin{algorithm}[H]
\textbf{Input:} training set $\mathcal{X}$, initial model $f\left(\cdot; \theta\right)$, removal threshold $\tau$, number of epochs $E$, batch size $M$, number of classes $K$, label assigned to ignored samples $i$ \\
\textbf{Output:} trained model $f\left(\cdot; \theta^*\right)$ \\
\For{epoch = 1,..., $E$}{
 \For{batch $\mathcal{B}$}{
  Sample a comparison batch $\widetilde{\mathcal{B}}$, $\widetilde{\mathcal{B}} \subset \mathcal{X}$, $|\widetilde{\mathcal{B}}| = M$\\
  Compute $\nabla\widetilde\mathcal{L}_{com}$ on $\widetilde{\mathcal{B}}$\\
  \For{$(x_t, y_t)$ $\in$ $\mathcal{B}$}{
  Compute $\nabla\widetilde\mathcal{L}\left(x_t,y_t\right)$ \\
  Set up corrected label vector $\widetilde{y}_t \leftarrow y_t$ \\
  \For{k=1, ..., $K$}{ \small
    $sim_{y_t}^k = sim\left(\left(\nabla\widetilde\mathcal{L}\left(x_t,y_t\right)\right)_{k}, \left(\nabla\widetilde\mathcal{L}_{com}\right)_{k}\right)$ (Eq.1) \\
    \If {$sim_{y_t}^k \leq 0$}{
        $\widetilde{y}_{t,k} \leftarrow i$}}
        $\mathcal{B} \leftarrow \mathcal{B} \setminus \{\left(x_t, y_t\right)\} \cup \{\left(x_t, \widetilde{y}_t\right)\}$
    
  }
  $\theta \leftarrow Optim\left(\theta, \mathcal{B}, \mathcal{L}\right)$
 }
 }
\caption{AGRA Algorithm for Multi-Label Datasets}
\label{alg:alg_multi}
\end{algorithm}

\vfill\null

\section{Dataset Preprocessing \& Statistics}

We used already preprocessed versions of text datasets For Chexpertin the Wrench framework \cite{wrench}.
For the medical imaging CheXpert dataset, where each class corresponds to one pathology that might be present in a training sample, we needed to additionally adapt the labels so that they can be used for our experiments.
In the original dataset a class value for each sample equals either -1 (uncertain), blank (not mentioned), 0 (negative), or 1 (positive). 
We binarized the label vector by mapping -1 to 1 (potentially increasing the noise ratio, but keeping more positive training samples), and blank to 0. 
The assignment of the value 0 to the blank labels is somewhat intuitive: not every radiographic report will mention all 12 pathologies, and if an observation is not even mentioned in the study report, there likely are no indications for it in the corresponding images \cite{garbin2021structured,irvin2019chexpert}.
Among the possible ways of handling the uncertainty labels (turning them to either 0 or 1, ignoring them, or keeping them as a separate class), we decided to choose the reassignment to label 1 as it was the best performing method for most of the CheXpert classes in \cite{irvin2019chexpert}. 
We focus on predicting the 12 pathologies in the CheXpert dataset, omitting the additional ``Support Devices'' and ``No Finding'' labels.

Table \ref{dataset_stat} provides information about the number of samples belonging to each class for all datasets used in the experiments.
Note that in a multi-class multi-label CheXpert dataset, one sample can belong to multiple classes.

\section{Data Encoding}

For the text datasets, we used feature-based TF-IDF vector representations.
The samples were encoded with the \texttt{TfidfVectorizer} tool available in Sklearn library\footnote{https://scikit-learn.org/stable/}.

For the CheXpert dataset we used fine-tuned convolutional neural networks (CNNs) as feature extractors 
following Giacomello et al. \cite{giacomello2021image}. 
In our experiments, the data samples were encoded using a fine-tuned EfficientNet-B0 \cite{tan2019efficientnet}.
The \texttt{torchvision} implementation \cite{torchvision} of EfficientNet-B0 used for our experiments was pre-trained on ImageNet. 
Despite the fact that the images contained in ImageNet greatly differ from chest radiographs, the pre-training still gives good intialization as was shown by Ke et al. \cite{ke2021chextransfer}.
The chest radiographs images were rescaled to a size of 224x224 and normalized with the mean and standard deviation from ImageNet. 
Then, the EfficientNet was fine-tuned on the training data for two epochs using Adam optimizer with learning rate 0.0001 and batch size 16.
Once the fine-tuning was completed, 1280-dimensional feature vectors for each image were extracted from the penultimate layer of the network and used as the data representations.

For CIFAR-10, we fine-tune a pre-trained ResNet-50 for 100 epochs with learning rate 0.001, batch size 64 and no weight decay and retrieve emeddings from the penultimate layer.
\newpage
\begin{table}[H]
\centering
\begin{tabular}{l|l|c}
\hline
Dataset                  & Class         & \%Samples \\
\hline
\hline
\multirow{2}{*}{YouTube} & SPAM          & $52.3\%$      \\
                         \cline{2-3}
                         & HAM           & $47.7\%$      \\
\hline
\multirow{2}{*}{SMS}     & SPAM          & $63.8\%$      \\
                         \cline{2-3}
                         & HAM           & $36.2\%$      \\
\hline
\multirow{6}{*}{TREC}    & DESC          & $56.6\%$     \\
                         \cline{2-3}
                         & ENTY          & $8.0\%$      \\
                         \cline{2-3}
                         & HUM           & $12.8\%$      \\
                         \cline{2-3}
                         & ABBR          & $1.0\%$       \\
                         \cline{2-3}
                         & LOC           & $8.0\%$      \\
                         \cline{2-3}
                         & NUM           & $13.6\%$      \\
\hline
\multirow{7}{*}{Yorùb\'a}& Africa        & $6.1\%$      \\
                         \cline{2-3}
                         & Entertainment & $25.2\%$      \\
                         \cline{2-3}
                         & Health        & $6.5\%$       \\
                         \cline{2-3}
                         & Nigeria       & $10.3\%$     \\
                         \cline{2-3}
                         & Politics      & $26.2\%$      \\
                         \cline{2-3}
                         & Sport         & $6.8\%$       \\
                         \cline{2-3}
                         & World         & $18.9\%$     \\
\hline
\multirow{5}{*}{Hausa}   & Africa        & $19.6\%$      \\
                         \cline{2-3}
                         & Health        & $10.1\%$      \\
                         \cline{2-3}
                         & Nigeria       & $8.6\%$       \\
                         \cline{2-3}
                         & Politics      & $7.9\%$      \\
                         \cline{2-3}
                         & World         & $53.7\%$      \\
\hline
\multirow{13}{*}{CheXpert} & Enlarged & \multirow{2}{*}{$10.4\%$}      \\
                         & Cardiomediastinum & \\
                         \cline{2-3}
                         & Cardiomegaly        & $15.8\%$      \\
                         \cline{2-3}
                         & Lung Opacity       & $49.8\%$       \\
                         \cline{2-3}
                         & Lung Lesion      & $4.8\%$      \\
                         \cline{2-3}
                         & Edema         & $29.1\%$      \\
                         \cline{2-3}
                         & Consolidation         & $19.1\%$      \\
                         \cline{2-3}
                         & Pneumonia         & $11.1\%$      \\
                         \cline{2-3}
                         & Atelectasis         & $30.1\%$      \\
                         \cline{2-3}
                         & Pneumothorax         & $10.1\%$      \\
                         \cline{2-3}
                         & Pleural Effusion         & $43.7\%$      \\
                         \cline{2-3}
                         & Pleural Other         & $2,8\%$      \\
                         \cline{2-3}
                         & Fracture         & $4.3\%$      \\
\hline
\end{tabular}
\caption{Percentage of samples belonging to each class in all data sets used for the experiments. Note that CheXpert dataset in designed for a multi-label classification, meaning more than one class can be assigned to one sample.}
\label{dataset_stat}
\end{table}


\section{The Hyper-Parameters and Search Space}
\label{sec:params_search_space}
\label{sec:params}

The hyper-parameters for our algorithm and baselines were retrieved with a grid search on the validation sets; each trial was performed 3 times with different initialization. 
The search space is provided in Table \ref{table:params}, and the exact selected values for our method AGRA can be found in Table \ref{table:selected_params}.
All AGRA results presented in the paper are reproducible with the seed value 0. 

\begin{table}[h]
\captionsetup{singlelinecheck=off}
\centering
\begin{tabular}{l| >{\centering\arraybackslash}p{1.3cm} | >{\centering\arraybackslash}p{1cm} | >{\centering\arraybackslash}p{1.3cm}}
\hline
Dataset &  Learning Rate $\eta$ & Batch Size  $M$ & Weight Decay $\lambda$\\
\hline
\hline
YouTube       & $1e-2$ & $32$  &  $1e-3$  \\
SMS           & $1e-3$ & $128$ &  $1e-3$  \\
TREC          & $1e-1$ & $32$  &  $1e-4$  \\
Yorùb\'a      & $1e-1$ & $512$ &  $1e-4$  \\
Hausa         & $1e-1$ & $512$ &  $1e-4$  \\
CIFAR-10      & $1e-3$ & $512$ &  $1e-4$  \\
CheXpert      & $1e-3$ & $128$ &  $1e-3$  \\
\hline
\end{tabular}
\caption[]{Selected hyper-parameters with grid search for the best AGRA configurations, which are:
    \begin{itemize}{
        \item for YouTube: $F1$ comparison loss, no weighted sampling, no alternative label,
        \item for SMS: $CE$ comparison loss, weighted sampling, no alternative label,
        \item for TREC: $F1$ comparison loss, weighted sampling,
        \item for Yorùb\'a: $F1$ comparison loss, weighted sampling,
        \item for Hausa: $F1$ comparison loss, weighted sampling,
        \item for CIFAR-10: $F1$ comparison loss, weighted sampling,
        \item for CheXpert: $F1$ comparison loss, weighted sampling not implemented.}
    \end{itemize}
}
\label{table:selected_params}
\end{table}

\begin{table}[h!]
\centering 
\begin{tabular}{ >{\centering\arraybackslash}p{1.8cm}| >{\centering\arraybackslash}p{2.2cm}|p{2.9cm}}
Method &  Hyper-parameter & Search Space \\
\hline
\hline
\multirow{4}{*}{AGRA}   &  learning rate $\eta$     & $1e-05$, $1e-04$, $1e-03$, $1e-02$, $1e-01$ \\
                        \cline{2-3}
                        &  batch size  $M$          &  $32$, $128$, $512$ \\
                        \cline{2-3}
                        &  weight decay $\lambda$   &  $1e-05$, $1e-04$, $1e-03$, $1e-02$, $1e-01$ \\
\hline
\multirow{3}{*}{No denoising}     &  learning rate $\eta$     &  $1e-05$, $1e-04$, $1e-03$, $1e-02$, $1e-01$ \\
                        \cline{2-3}
                        &  batch size  $M$          &  $32$, $64$, $128$ \\
                        \cline{2-3}
                        &  weight decay $\lambda$   &  $1e-05$, $1e-04$, $1e-03$, $1e-02$, $1e-01$ \\
\hline
\multirow{3}{*}{DP}     &  DP learning rate         &  $1e-05$, $1e-04$, $1e-03$, $1e-02$, $1e-01$ \\
                        \cline{2-3}
                        &  DP batch size            &  $32$, $64$, $128$ \\
                        \cline{2-3}
                        &  DP num epochs             &  $5$, $10$, $50$, $100$, $200$ \\
                        \cline{2-3}
                        &  learning rate $\eta$     &  $1e-05$, $1e-04$, $1e-03$, $1e-02$, $1e-01$ \\
                        \cline{2-3}
                        &  batch size  $M$          &  $32$, $64$, $128$ \\
                        \cline{2-3}
                        &  weight decay $\lambda$   &  $1e-05$, $1e-04$, $1e-03$, $1e-02$, $1e-01$ \\
\hline
\multirow{3}{*}{MeTal}  &  MeTal learning rate      &  $1e-05$, $1e-04$, $1e-03$, $1e-02$, $1e-01$ \\
                        \cline{2-3}
                        &  MeTal weight decay       &  $32$, $64$, $128$ \\
                        \cline{2-3}
                        &  MeTal num epochs         &  $5$, $10$, $50$, $100$, $200$ \\
                        \cline{2-3}
                        &  learning rate $\eta$     &  $1e-05$, $1e-04$, $1e-03$, $1e-02$, $1e-01$ \\
                        \cline{2-3}
                        &  batch size  $M$          &  $32$, $64$, $128$ \\
                        \cline{2-3}
                        &  weight decay $\lambda$   &  $1e-05$, $1e-04$, $1e-03$, $1e-02$, $1e-01$ \\
\hline
\multirow{3}{*}{FlyingSquid}     &  learning rate $\eta$     &  $1e-05$, $1e-04$, $1e-03$, $1e-02$, $1e-01$ \\
                        \cline{2-3}
                        &  batch size  $M$          &  $32$, $64$, $128$ \\
                        \cline{2-3}
                        &  weight decay $\lambda$   &  $1e-05$, $1e-04$, $1e-03$, $1e-02$, $1e-01$ \\
\hline
\multirow{4}{*}{Cleanlab}     &  CL n\_folds              &  $2$, $5$, $7$, $9$, $12$ \\
\cline{2-3} & learning rate $\eta$     &  $1e-05$, $1e-04$, $1e-03$, $1e-02$, $1e-01$ \\
\cline{2-3} & batch size  $M$ &  $32$, $128$, $512$\\
\cline{2-3} &  weight decay $\lambda$ &  $1e-05$, $1e-04$, $1e-03$\\
\hline
\multirow{3}{*}{CORES$^2$}     & 
learning rate $\eta$ &  $1e-05$, $1e-04$, $1e-03$, $1e-02$, $1e-01$ \\
\cline{2-3} & batch size  $M$ &  $32$, $128$, $512$\\
\cline{2-3} &  weight decay $\lambda$ &  $1e-05$, $1e-04$, $1e-03$, $1e-02$, $1e-01$\\
\end{tabular}
\caption{Hyper-parameters and search space used for a grid search. 
The search spaces for DP, MeTaL, and FS methods were inherited from Wrench \cite{wrench}.
The search spaces for Cleanlab and AGRA were selected empirically.}
\label{table:params}
\end{table}

For CheXpert, the hyperparameters were fixed to learning rate $1e-3$, weight decay $1e-3$ and batch size 128 without grid search for AGRA and Cleanlab due to energy
considerations and resource constraints. 
The Cleanlab hyperparameter n\_folds was set to 2 for CheXpert.
For CORES$^2$, we choose $\beta=0.2\cdot K$, where $K$ is the number of classes, as suggested by \cite{cheng2020learning}. 
For CIFAR, we also tuned the decision threshold value as a hyper-parameter (the range: 0.001, 0.01, 0.5).
The best hyper-parameter values were selected based on one trial.







\end{document}